\documentclass[11pt,a4paper]{article}
\usepackage[hyperref]{eacl2021}
\usepackage{times}
\usepackage{latexsym}
\usepackage{url}
\usepackage{booktabs}
\usepackage{appendix}
\usepackage{xcolor}
\usepackage{algorithm}
\usepackage{algorithmic}
\usepackage{amsmath}
\usepackage{amsfonts}
\usepackage{natbib}
\usepackage{tabularx}
\usepackage{multirow}
\usepackage{tikz}
\usepackage{graphicx}
\usepackage[position=bottom]{subfig}

\usepackage{microtype}

\aclfinalcopy 
\def\aclpaperid{*} 

\definecolor{red}{HTML}{dc322f}

\newcommand{\opt}[1]{\textbf{#1}}

\newcommand{\advsubst}[6]{
$\overset{\raise0.0em\hbox{\colorbox{blue!#1}{#2}}}{\textbf{#3}}\!\!\overset{\raise.3em\hbox{$\dashleftarrow$\,\,}}{\textcolor{white}{I}}\!\!\overset{\colorbox{blue!#4}{#5}}{\text{[}\textcolor{red}{\textit{\textbf{#6}}}\text{]}}$
}

\newcommand{\advsubstleft}[6]{
$\overset{\raise0.0em\hbox{\colorbox{blue!#1}{#2}}}{\textbf{#3}}\!\!\overset{\raise.3em\hbox{$\,\dashleftarrow$}}{\textcolor{white}{I}}\!\!\overset{\colorbox{blue!#4}{#5}}{\text{[}\textcolor{red}{\textit{\textbf{#6}}}\text{]}}$
}

\newcommand{\advsubstnoc}[4]{
$\!\overset{#1}{\textbf{#2}}$ $\overset{#3}{\text{[}\textcolor{red}{\textit{\textbf{#4}}}\text{]}}\!$
}

\title{Frequency-Guided Word Substitutions for Detecting\\Textual Adversarial Examples}
 \author{
 Maximilian Mozes$^1$\quad Pontus Stenetorp$^1$\quad Bennett Kleinberg$^{2,1}$\quad Lewis D. Griffin$^1$\\
 $^1$University College London\\
 $^2$Tilburg University\\
 \small{\texttt{\{m.mozes, p.stenetorp, l.griffin\}@cs.ucl.ac.uk}}\\
 \small{\texttt{bennett.kleinberg@tilburguniversity.edu}}
 }

\begin{document}
\maketitle
\begin{abstract}
Recent efforts have shown that neural text processing models are vulnerable to adversarial examples, but the nature of these examples is poorly understood. In this work, we show that adversarial attacks against CNN, LSTM and Transformer-based classification models perform word substitutions that are identifiable through frequency differences between replaced words and their corresponding substitutions. Based on these findings, we propose \textit{frequency-guided word substitutions} (\textsc{FGWS}), a simple algorithm exploiting the frequency properties of adversarial word substitutions for the detection of adversarial examples. \textsc{FGWS} achieves strong performance by accurately detecting adversarial examples on the SST-2 and IMDb sentiment datasets, with $F_1$ detection scores of up to 91.4\% against RoBERTa-based classification models. We compare our approach against a recently proposed perturbation discrimination framework and show that we outperform it by up to 13.0\% $F_1$.
\end{abstract}

\section{Introduction}
Artificial neural networks are vulnerable to adversarial examples---carefully crafted perturbations of input data that lead a learning model into making false predictions~\cite{szegedy2014}.

\begin{figure}[t]
\centering
\resizebox{1.0\columnwidth}{!}{
\begin{tabular}{cl} 
\toprule
\multicolumn{1}{c}{\textbf{Attack}} & \multicolumn{1}{l}{\textbf{Original or perturbed sequence}} \\\midrule
  \multicolumn{1}{c}{None} & A clever blend of fact and fiction \\\\
  \textsc{Genetic} & A \advsubstleft{15}{1.39}{brainy}{40}{5.55}{clever} blend of fact and fiction \\\\
  \textsc{PWWS} & A \advsubst{17}{1.61}{cunning}{40}{5.55}{clever} \advsubst{10}{0.00}{blending}{30}{3.81}{blend} of\\& fact and \advsubst{10}{0.00}{fabrication}{35}{4.39}{fiction} \\ \bottomrule
\end{tabular}
}
\caption{Corpus $\log_e$ frequencies of the replaced words (bold, italic, red) and their corresponding adversarial substitutions (bold, black) using the \textsc{Genetic}~\cite{alzantot-etal-2018-generating} and \textsc{PWWS}~\cite{ren-etal-2019-generating} attacks on SST-2~\cite{socher-etal-2013-recursive}.}
\label{fig:attack-examples}
\end{figure}
While initially discovered for computer vision tasks, natural language processing (NLP) models have also been shown to be oversensitive to adversarial input perturbations for a variety of tasks~\cite{Papernot2016CraftingAI,jia-liang-2017-adversarial,belinkov2017synthetic,glockner-etal-2018-breaking, iyyer-etal-2018-adversarial}. Here we focus on highly successful synonym substitution attacks~\cite{alzantot-etal-2018-generating, ren-etal-2019-generating, zang-etal-2020-word}, in which individual words are replaced with semantically similar ones. Existing defense methods against these attacks mainly focus on adversarial training~\cite{jia-liang-2017-adversarial, ebrahimi2018, ribeiro-etal-2018-semantically, ren-etal-2019-generating, jin2019bert} and hence typically require a priori attack knowledge and models to be retrained from scratch to increase their robustness. Recent work by~\citet{zhou-etal-2019-learning} instead proposes \textsc{DISP} (\textit{learning to \underline{dis}criminate \underline{p}erturbations}), a perturbation discrimination framework that exploits pre-trained contextualized word representations to detect and correct word-level adversarial substitutions without having to retrain the attacked model. In this paper, we show that we can achieve an improved performance for the detection and correction of adversarial examples based on the finding that various word-level adversarial attacks have a tendency to replace input words with less frequent ones.\footnote{This frequency difference is expected for attacks that explicitly conduct symbol substitutions resulting in out-of-vocabulary (OOV) terms~\cite{2018arXiv180104354G}. We therefore study attacks that do not explicitly enforce a mapping to words that have lower frequencies.} Figure~\ref{fig:attack-examples} illustrates this tendency for two state-of-the-art attacks. We provide statistical evidence to support this observation and propose a rule-based and model-agnostic algorithm, \textit{frequency-guided word substitutions} (FGWS), to detect adversarial sequences and recover model performances for perturbed test set sequences. \textsc{FGWS} effectively detects adversarial perturbations, achieving $F_1$ scores of up to 91.4\% against RoBERTa-based models~\cite{liu2019roberta} on the IMDb sentiment dataset~\cite{Maas2011}. Furthermore, our results show that \textsc{FGWS} outperforms \textsc{DISP} by up to 13.0\% $F_1$ when differentiating between unperturbed and perturbed sequences, despite representing a conceptually simpler approach to this task.

\section{Generating adversarial examples}
\label{sec:gen-adv-examples}
In our experiments, we investigate two baseline attacks introduced by~\citet{ren-etal-2019-generating} as well as two state-of-the-art attacks.

\textbf{\textsc{Random}}.
Our first baseline attack is a simple word substitution model that randomly selects words in an input sequence and replaces them with synonyms randomly sampled from a set of synonyms related to the specific word. We follow~\citet{ren-etal-2019-generating} by using \textsc{WordNet}~\cite{fellbaum98wordnet} to identify synonym substitutions for each selected word. 

\textbf{\textsc{Prioritized}}.
Our second baseline builds upon \textsc{Random} by selecting the replacement word from the synonym set that maximizes the change in prediction confidence for the true label of an input. 

\textbf{\textsc{Genetic}}.
We additionally analyze an attack suggested by~\citet{alzantot-etal-2018-generating}, consisting of a population-based black-box mechanism based on genetic search that iteratively performs individual word-level perturbations to an input sequence to cause a misclassification. 

\textbf{\textsc{PWWS}}. Lastly, we analyze the \textit{probability weighted word saliency} (\textsc{PWWS}) algorithm~\cite{ren-etal-2019-generating}. For each word in an input sequence, \textsc{PWWS} selects a set of synonym replacements from \textsc{WordNet} and chooses the synonym yielding the highest difference in prediction confidence for the true class label after replacement. The algorithm furthermore computes the word saliency~\cite{li-etal-2016-visualizing, word-saliency-li} for each input word and ranks word replacements based on these two indicators. 

\paragraph{Datasets and models.}We perform experiments on two binary sentiment classification datasets, the \textit{Stanford Sentiment Treebank}~\cite[SST-2,][]{socher-etal-2013-recursive} and the IMDb reviews dataset~\cite{Maas2011}, both of which are widely used in related work focusing on adversarial examples in NLP~\cite{jia2019certified, ren-etal-2019-generating,zhou-etal-2019-learning}. Dataset details can be found in Appendix~\ref{sec:app-datasets}. Adhering to~\citet{zhou-etal-2019-learning}, we attack a pre-trained model based on the Transformer architecture~\cite{vaswani2017}. \citet{zhou-etal-2019-learning} use \textsc{BERT}~\cite{devlin-etal-2019-bert} in their experiments, but we found that RoBERTa~\cite{liu2019roberta} represents a stronger model for the specified tasks. 

We additionally experiment with both a CNN~\cite{kim2014convolutional} and an LSTM~\cite{hochreiter97} text classification model, both of which have been employed in existing work studying textual adversarial attacks~\cite{alzantot-etal-2018-generating, lei2019discrete, jia2019certified, tsai-etal-2019-adversarial-attack, ren-etal-2019-generating}. 

The fine-tuned RoBERTa model achieves 93.4\% and 94.9\% accuracy on the IMDb and SST-2 test sets, which is comparable to existing work~\cite{beltagy2020longformer, liu2019roberta}. On the IMDb test set, the CNN achieves an accuracy of 86.0\% and the LSTM achieves 83.1\%. These performances are close to existing work using comparable settings~\cite{zhang-etal-2019-generating-fluent, ren-etal-2019-generating}. On the SST-2 test set, the CNN achieves 84.0\% and the LSTM 85.2\% accuracy, which are also close to comparable experiments~\cite{huang-etal-2019-achieving}.

Following~\citet{ren-etal-2019-generating}, we apply all four attacks to a random subset of 2,000 sequences from the IMDb test set as well as the entire test set of SST-2 (1,821 samples). Implementation details for the models and attacks can be found in Appendix~\ref{sec:app-attack-details}. We report the after-attack accuracies\footnote{The after-attack accuracy represents the model accuracy on the test set after perturbing all correctly classified inputs. A lower after-attack accuracy indicates a stronger attack.} for the RoBERTa model in Table~\ref{tab:disp-comparison} and for the CNN/LSTM models in Table~\ref{tab:detection-performance} (column Adv.). We observe that all four attacks cause notable decreases in model accuracy on the test sets, and that \textsc{Genetic} and \textsc{PWWS} are more successful than the baseline attacks in most comparisons.

\section{Analyzing frequencies of adversarial word substitutions}
\label{sec:statistical-analysis}
Next, we conduct an analysis of the word frequencies of individual words replaced by the attacks and their substitutions. We compute the $\log_e$ training set frequencies $\phi(x)$ of all words $x$ that have been replaced by the respective attacks and all of their corresponding substitutions. Then, we conduct Bayesian hypothesis testing~\cite{rouder2009bayesian} to statistically compare the two samples. This is achieved by computing the Bayes factor $\mathrm{BF}_{10}$, representing the degree to which the data favor the alternative hypothesis over the null hypothesis. Here, the alternative hypothesis $\mathcal{H}_1$ states that \textit{the frequencies of replaced words differ from the frequencies of the adversarial substitutions}. The null hypothesis $\mathcal{H}_0$ states that \textit{there is no such difference}. The higher $\mathrm{BF}_{10}$, the stronger the evidence in favor of the alternative hypothesis $\mathcal{H}_1$.\footnote{A Bayes factor $\mathrm{BF}_{10}> 100$ can be interpreted as 
``extreme" evidence for  $\mathcal{H}_1$~\cite{wagenmakers2011psychologists}.} We additionally calculate Cohen's $d$ effect sizes for all mean frequency comparisons.\footnote{Cohen's $d$ indicates the magnitude of the frequency differences of the two samples---larger effect sizes suggest a higher magnitude of the frequency difference. A value of $d = 0.8$ can be interpreted as a large effect, $d=0.5$ is considered a moderate effect~\cite{cohen2013statistical}.}

\begin{table}[t]
\centering
\resizebox{\columnwidth}{!}{
\begin{tabular}{clcccccccc}
\toprule
\multirow{2}[2]{*}{\textbf{Dataset}} & \multicolumn{1}{c}{\multirow{2}[2]{*}{\textbf{Attack}}} & \multicolumn{2}{c}{\textbf{Replaced}} & \multicolumn{3}{c}{\textbf{Subst.}} &
\multicolumn{3}{c}{\textbf{non-OOV}} \\
\cmidrule(lr){3-4}\cmidrule(lr){5-7}\cmidrule(l){8-10}
& & $\mu_{\phi}$ & $\sigma_{\phi}$ & $\mu_{\phi}$ & $\sigma_{\phi}$ & $d$ & $\mu_{\phi}$ & $\sigma_{\phi}$ & $d$ \\
\midrule
\multirow{4}[1]{*}{IMDb} & \textsc{Random} & 7.6 & 2.5 & 3.4 & 2.8 & \opt{1.6} & 4.4 & 2.4 & 1.3 \\
 &  \textsc{Prioritized} & 7.6 & 2.5 & 3.6 & 2.8 & 1.5 & 4.4 & 2.4 & 1.3 \\
 &  \textsc{Genetic} & 6.5 & 2.0 & 3.7 & 2.3 & 1.3 & 4.0 & 2.2 & 1.2 \\
 &  \textsc{PWWS} & 6.9 & 2.3 & 4.4 & 2.5 & 1.0 & 5.0 & 2.1 & 0.9 \\ \midrule
\multirow{4}[1]{*}{SST-2} & \textsc{Random} & 5.4 & 2.6 & 2.1 & 2.4 & \opt{1.4} & 4.0 & 1.8 & 0.6 \\
 &  \textsc{Prioritized} & 5.4 & 2.6 & 2.1 & 2.4 & 1.3 & 4.0 & 1.8 & 0.6 \\
 &  \textsc{Genetic} & 4.4 & 1.9 & 1.9 & 2.2 & 1.2 & 3.6 & 1.6 & 0.4 \\
 &  \textsc{PWWS} & 4.8 & 2.1 & 2.9 & 2.2 & 0.9 & 4.0 & 1.5 & 0.4 \\
 \bottomrule
\end{tabular}
}
\caption{Mean $\log_e$ frequencies of replaced words and their substitutions. Values in bold denote largest effect sizes per dataset.}
\label{tab:statistical-results}
\end{table}

Table~\ref{tab:statistical-results} shows the $\log_e$ frequencies (mean $\mu_{\phi}$ and standard deviation $\sigma_{\phi}$) and Cohen's $d$ for the specified samples generated by the attacks against the RoBERTa model (the results for the CNN and LSTM models can be found in Appendix~\ref{sec:app-freq-diff-cnn-lstm}). We report the mean frequencies of all adversarial substitutions (Subst.) and only those that occur in the training set (non-OOV), to demonstrate that the frequency differences are not solely caused by OOV substitutions. Across datasets and attacks, the substitutions are consistently less frequent than the words selected for replacement. We observe large Cohen's $d$ effect sizes for the majority of comparisons, statistically supporting the observation of mean frequency differences between replaced words and their corresponding substitutions. We furthermore observe that $\mathrm{BF}_{10} > 10^{55}$ holds for all comparisons---both when considering all and only non-OOV substitutions (the $\mathrm{BF}_{10}$ scores can be found in Appendix~\ref{sec:app-bayes-factor}). This provides strong empirical evidence that $\mathcal{H}_1$ is more likely to be supported by the measured word frequencies (see Appendix~\ref{sec:app-freq-histograms} for additional illustrations).

\begin{table*}[t]
\centering
\resizebox{0.8\textwidth}{!}{
\begin{tabular}{clccccccc}
\toprule
 \multirow{2}[2]{*}{\textbf{Dataset}} & \multicolumn{1}{c}{\multirow{2}[2]{*}{\textbf{Attack}}} & \multicolumn{1}{c}{\multirow{2}[2]{*}{\textbf{Adv.}}} & \multicolumn{2}{c}{\textbf{Restored acc.}}  & \multicolumn{2}{c}{\textbf{TPR} (\textbf{FPR})} & \multicolumn{2}{c}{$F_1$} \\ \cmidrule(lr){4-5} \cmidrule(lr){6-7} \cmidrule(l){8-9}  & & & \multicolumn{1}{c}{\textsc{DISP}} & \multicolumn{1}{c}{\textsc{FGWS}} & \multicolumn{1}{c}{\textsc{DISP}} & \multicolumn{1}{c}{\textsc{FGWS}} & \multicolumn{1}{c}{\textsc{DISP}} & \multicolumn{1}{c}{\textsc{FGWS}} \\ \midrule
\multicolumn{1}{c}{\multirow{4}{*}{IMDb}} & \textsc{Random} & 87.3 & 89.2 & \opt{91.0} & 63.6 (9.4) & \opt{83.5} (9.3) & 73.6 & \opt{86.6} \\
\multicolumn{1}{c}{} &  \textsc{Prioritized} & 41.5 & 81.0 & \opt{85.9} & 87.8 (9.4) & \opt{92.0} (9.3) & 89.0 & \opt{91.4} \\
\multicolumn{1}{c}{} & \textsc{Genetic} & 47.7 & 74.1 & \opt{80.6} & 70.4 (9.4) & \opt{81.5} (9.3) & 78.3 & \opt{85.4} \\
\multicolumn{1}{c}{} & \textsc{PWWS} & 41.0 & 68.7 & \opt{75.4} & 66.2 (9.4) & \opt{76.4} (9.3) & 75.4 & \opt{82.3} \\ \midrule
\multicolumn{1}{c}{\multirow{4}{*}{SST-2}}  & \textsc{Random} & 87.2 & 86.6 & \opt{90.0} & \opt{66.2} (11.9) & 61.3 (11.4) & \opt{74.4} & 71.0 \\
\multicolumn{1}{c}{} & \textsc{Prioritized} & 68.9 & 80.8 & \opt{84.8} & 69.1 (11.9) & \opt{74.7} (11.4) & 76.3 & \opt{80.3} \\
\multicolumn{1}{c}{} &  \textsc{Genetic} & 40.8 & 60.1 & \opt{61.7} & \opt{57.2} (11.9) & 57.0 (11.4) & \opt{67.7} & \opt{67.7} \\
\multicolumn{1}{c}{} &  \textsc{PWWS} & 57.4 & 71.0 & \opt{78.2} & 59.6 (11.9) & \opt{65.6} (11.4) & 69.6 & \opt{74.2} \\ \bottomrule
\end{tabular}
}
\caption{Adversarial example detection performances for \textsc{DISP} and \textsc{FGWS} when evaluated on attacks against RoBERTa. Adv. shows the model's classification accuracy on the perturbed sequences. Restored acc. denotes model accuracy on the adversarial sequences after transformation. Values in bold represent best scores per metric, dataset and attack.}
\label{tab:disp-comparison}
\end{table*}

\begin{table}[t]
\centering
\resizebox{\columnwidth}{!}{
\begin{tabular}{clccccc}
\toprule
 \multirow{2}[2]{*}{\shortstack{\textbf{Model}/\\\textbf{Dataset}}} & \multicolumn{1}{c}{\multirow{2}[2]{*}{\textbf{Attack}}} & \multicolumn{1}{c}{\multirow{2}[2]{*}{\textbf{Adv.}}} & \multicolumn{2}{c}{\textbf{Restored acc.}} & \multicolumn{2}{c}{\textbf{$F_1$}} \\ \cmidrule(lr){4-5} \cmidrule(l){6-7}
 & & & \textsc{NWS} & \textsc{FGWS} & \textsc{NWS} & \textsc{FGWS} \\ \midrule
 \multirow{4}{*}{\shortstack{CNN/\\IMDb}} 
 & \textsc{Random} & 73.0 & 79.5 & \opt{84.7} & 75.2 & \opt{83.5} \\
 & \textsc{Prioritized} & 14.0 & 41.6 & \opt{78.9} & 71.5 & \opt{89.3} \\
 & \textsc{Genetic} & 10.7 & 21.3 & \opt{68.5} & 37.9 & \opt{83.5} \\
 & \textsc{PWWS} & 10.2 & 27.4 & \opt{70.2} & 45.4 & \opt{83.9} \\ \midrule
\multirow{4}{*}{\shortstack{LSTM/\\IMDb}} 
& \textsc{Random} & 64.7 & 75.7 & \opt{80.9} & 80.5 & \opt{83.9} \\
& \textsc{Prioritized} & 3.2 & 32.0 & \opt{71.6} & 62.4 & \opt{86.6} \\
& \textsc{Genetic} & 1.2 & 10.9 & \opt{54.9} & 34.3 & \opt{78.0} \\
& \textsc{PWWS} & 1.6 & 17.3 & \opt{57.1} & 41.7 & \opt{77.4} \\ \midrule
\multirow{4}{*}{\shortstack{CNN/\\SST-2}} 
& \textsc{Random} & 71.8 & 77.1 & \opt{78.4} & \opt{71.4} & 69.2 \\
& \textsc{Prioritized} & 50.3 & 60.1 & \opt{69.3} & 54.8 & \opt{67.8} \\
& \textsc{Genetic} & 19.6 & 34.9 & \opt{48.8} & 49.9 & \opt{60.3} \\
& \textsc{PWWS} & 28.1 & 47.4 & \opt{58.1} & 55.1 & \opt{63.9} \\ \midrule
\multirow{4}{*}{\shortstack{LSTM/\\SST-2}} 
& \textsc{Random} & 73.4 & 79.3 & \opt{80.5} & \opt{69.2} & 62.2 \\
& \textsc{Prioritized} & 48.5 & 59.9 & \opt{74.0} & 54.9 & \opt{67.3} \\
& \textsc{Genetic} & 21.3 & 37.6 & \opt{61.1} & 51.2 & \opt{62.8} \\
& \textsc{PWWS} & 28.6 & 49.7 & \opt{67.2} & 55.9 & \opt{63.4} \\ \bottomrule
\end{tabular}
}
\caption{Performance results of \textsc{NWS} and \textsc{FGWS} on attacks against the CNN and LSTM models. Values in bold indicate best performances per model-dataset-attack combination and metric.}
\label{tab:detection-performance}
\end{table}

\section{Frequency-guided word substitutions}
\label{sec:detection-mechanism-section}
Based on the observation of consistent frequency differences between replaced words and adversarial substitutions, we argue that the effects of such substitutions can be mitigated through simple frequency-based transformations. To do this, we propose \textit{frequency-guided word substitutions} (\textsc{FGWS}), a detection method that estimates whether a given input sequence is an adversarial example.\footnote{Code is available at \url{https://github.com/maximilianmozes/fgws}.} We denote a classification model by a function $f(X)$ that maps a sequence $X$ to a $C$-dimensional vector representing the probabilities for predicting each of the $C$ possible classes. We represent a sequence as $X =\{x_1,\dots,x_n\}$, where $x_i$ denotes the $i$-th word in the sequence. We furthermore introduce the notation $f^\ast(X) \in \{1,\dots, C\}$ representing the class label predicted by $f$ given input $X$. \textsc{FGWS} transforms a given sequence $X$ into a sequence $X^\prime$ by replacing infrequent words with more frequent, semantically similar substitutions. 
We initially define the subset $X_E := \{x \in X \, |\, \phi(x) < \delta\}$ of words that are eligible for substitution, where $\delta \in \mathbb{R}_{>0}$ is a frequency threshold. \textsc{FGWS} then generates a sequence $X^\prime$ from $X$ by replacing all eligible words with words that are semantically similar, but have higher occurrence frequencies in the model's training corpus. For each eligible word $x\in X_E$ we consider the set of replacement candidates $\mathcal{S}(x)$ and find a replacement $x^\prime$ by selecting $x^\prime = {\mathrm{argmax}}_{w \in \mathcal{S}(x)} \,\, \phi(w)$. We then generate $X^\prime$ by replacing each eligible word $x$ with $x^\prime$ if $\phi(x^\prime) > \phi(x)$. Given the prediction label $y = f^\ast(X)$ for $X$ and a threshold $\gamma\in[0,1]$, the sequence $X$ is considered adversarial if $f(X)_y - f(X^\prime)_y > \gamma$, i.e., if the difference in prediction confidence on class $y$ before and after transformation exceeds the threshold $\gamma$. The threshold allows control of the rate of false positives (i.e., unperturbed sequences that are erroneously identified as adversarial) flagged by our method.

\subsection{Comparisons}

\paragraph{\textsc{DISP}.}We compare \textsc{FGWS} to the \textsc{DISP} framework~\cite{zhou-etal-2019-learning}, which is, to the best of our knowledge, the best existing approach for the detection of word-level adversarial examples. \textsc{DISP} uses two independent BERT-based components, a perturbation discriminator and an embedding estimator for token recovery, to identify perturbed tokens and to reconstruct the replaced ones.

\paragraph{\textsc{NWS}.}For the CNN and LSTM models, we compare \textsc{FGWS} with the \textit{naive word substitutions} (\textsc{NWS}) baseline. For a given input sequence, \textsc{NWS} selects all OOV words in that sequence and replaces each with a random choice from a set of semantically related words. We restrict \textsc{NWS} to allow only substitutions for which the replacement word occurs in the model's training vocabulary. \textsc{NWS} can be interpreted as a variant of \textsc{FGWS} that is not explicitly guided by word frequencies.

\subsection{Experiments}
\label{subsec:experiments}
We apply both methods to the adversarial examples crafted by the four attacks on the subsets of both the IMDb and SST-2 datasets as described in Section~\ref{sec:gen-adv-examples}. To account for an imbalance between unperturbed and perturbed sequences, we repeatedly bootstrap a balanced set of unperturbed sequences for each set of perturbed sequences for 10,000 times and compute the average detection scores. For \textsc{FGWS}, we tune the frequency threshold $\delta$ for each model-dataset combination on the validation set. To do this, we utilize the \textsc{Prioritized} attack to craft adversarial examples from all sequences of the validation set\footnote{We assume both baseline attacks as given to the defender, and prefer \textsc{Prioritized} over \textsc{Random} due to increased effectiveness and hence a larger sample size for parameter tuning.} and compare \textsc{FGWS} detection performances with different values for $\delta$. Specifically, we set $\delta$ equal to the $\log_e$ frequency representing the $q^{\text{th}}$ percentile of all $\log_e$ frequencies observed by the words eligible for replacement in the training set, and experiment with $q \in \{0, 10, \dots, 100\}$. We select $\gamma$ so that not more than $10\%$ of the unperturbed sequences in the validation set are labeled as adversarial.\footnote{We provide additional results with varying false positive thresholds in Appendix~\ref{sec:app-varying-fpr}.} For \textsc{FGWS}, we define the set of replacement candidates for each word $x\in X_E$ as the union of the word's $K$ nearest neighbors in a pre-trained \textsc{GloVe}~\cite{pennington-etal-2014-glove} word embedding space and its synonyms in \textsc{WordNet}. We set $K$ equal to the average number of \textsc{WordNet} synonyms for each word in the validation set (yielding $K=6$ for IMDb and $K=8$ for SST-2).

\begin{figure*}[t]
\centering
\resizebox{0.9\textwidth}{!}{
\begin{tikzpicture}
\node (table) [inner sep=0pt] {
\begin{tabular}{lllr}
\multicolumn{2}{l}{\underline{Unperturbed}} & a smart sweet and playful romantic comedy & \textit{positive} (99.9\%) \\
(A) & \textsc{PWWS} & a \advsubstnoc{0.00}{impertinent}{5.69}{smart} \advsubstnoc{1.79}{odoriferous}{5.77}{sweet} and playful romantic comedy & \textit{negative} (56.3\%) \\
 (D) & \textsc{DISP} & \advsubstnoc{10.22}{the}{9.99}{a} \advsubstnoc{6.83}{little}{0.00}{impertinent} odoriferous and playful romantic comedy & \textit{positive} (79.3\%) \\
 (D) & \textsc{FGWS} & a \advsubstnoc{5.69}{smart}{0.00}{impertinent} \advsubstnoc{5.77}{sweet}{1.79}{odoriferous} and playful romantic comedy & \textit{positive} (99.9\%) \\
\end{tabular}
};
\draw [rounded corners=.5em] (table.north west) rectangle (table.south east);
\end{tikzpicture}
}
\caption{The detection methods applied to an adversarial example from the \textsc{PWWS} attack against RoBERTa on SST-2. The words highlighted in bold, italic and red were selected for replacement by the attack (A) and the detection methods (D), the ones in bold and black denote the substitutions. The values above the words denote their $\log_e$ frequencies.}
\label{tab:detection-example}
\end{figure*}

\subsection{Results}
We report the results comparing \textsc{FGWS} to \textsc{DISP} on attacks against RoBERTa in Table~\ref{tab:disp-comparison}. Here, the true positive rate (TPR) represents the percentage of successful adversarial examples that were correctly identified as such, and the false positive rate (FPR) denotes the percentage of unperturbed sequences that were identified as adversarial. The column Adv. gives the classification accuracy on the perturbed sequences, and Restored acc. the model's accuracy on the adversarial sequences after transformation. We observe that \textsc{FGWS} best restores the model's classification accuracy across all comparisons, showing it to be effective in mitigating the effects of the individual attacks. Furthermore, \textsc{FGWS} outperforms \textsc{DISP} in terms of true positive rates and $F_1$ across the majority of experiments. These results show that, although contextualized word representations (\textsc{DISP}) serve as a competitive method to detect adversarial examples, relying solely on frequency-guided substitutions (\textsc{FGWS}) shows to be more effective. Figure~\ref{tab:detection-example} provides an example adversarial sequence generated with the \textsc{PWWS} attack and the two corresponding transformed sequences using \textsc{DISP} and \textsc{FGWS} (see Appendix~\ref{sec:app-fgws-examples} for additional examples). 

The results of \textsc{NWS} and \textsc{FGWS} against the CNN and LSTM models are shown in Table~\ref{tab:detection-performance}. We observe that \textsc{FGWS} outperforms \textsc{NWS} across all comparisons in terms of restored model accuracy and in the majority of comparisons in terms of $F_1$. Moreover, the direct comparison between \textsc{NWS} and \textsc{FGWS} again underlines the importance of utilizing word frequencies as guidance for the word substitutions: while \textsc{NWS} is not guided by word frequency characteristics to perform the word replacements, we observe that \textsc{FGWS} outperforms \textsc{NWS} by a large margin in most comparisons, demonstrating the effectiveness of mapping infrequent words to their most frequent semantically similar counterparts to detect adversarial examples.

\subsection{FGWS on unperturbed data}
We furthermore investigate the effect of \textsc{FGWS} on model performance on unperturbed sequences after transformation. To do this, we transform the sampled test sets using \textsc{FGWS} and evaluate classification accuracies after sequence transformation. The differences in accuracy for the CNN, LSTM and RoBERTa models before and after transformation are $0.0$\%, $+1.0$\% and $-0.2$\% for IMDb and $-1.8$\%, $-2.9$\% and $-1.8$\% for SST-2. This indicates that FGWS applied to unperturbed data has only small effects on classification accuracy, and in some cases even slightly increases prediction accuracy.

\section{Limitations}
It is worth mentioning that compared to \textsc{FGWS}, \textsc{DISP} represents a more general perturbation discrimination approach since it is trained to detect both character- and word-level adversarial perturbations, whereas \textsc{FGWS} solely focuses on word-level attacks.

Furthermore, it remains open whether \textsc{FGWS} would be effective against attacks for which the frequency difference is less evident. To investigate this, we conducted preliminary experiments by restricting the investigated attacks to only allow equifrequent substitutions. However, we observed that introducing this constraint has a substantial effect on attack performance, since the attacks are supplied with fewer candidate replacements. We will further investigate this in future work.

\section{Conclusion}
We have shown that the word frequency characteristics of adversarial word substitutions can be leveraged effectively to detect adversarial sequences for neural text classification. Our proposed approach outperforms existing detection methods despite representing a conceptually simpler approach to this task.

\section*{Acknowledgements}
This research was supported by the Dawes Centre for Future Crime at University College London.

\bibliography{eacl2021}
\bibliographystyle{acl_natbib}

\appendix

\begin{table*}[t]
\centering
\resizebox{0.75\textwidth}{!}{
\begin{tabular}{cclcccccccc}
\toprule
\multirow{2}[2]{*}{\textbf{Dataset}} & \multirow{2}[2]{*}{\textbf{Model}} & \multicolumn{1}{c}{\multirow{2}[2]{*}{\textbf{Attack}}} & \multicolumn{2}{c}{\textbf{Replaced}} & \multicolumn{3}{c}{\textbf{Subst.}} &
\multicolumn{3}{c}{\textbf{non-OOV}} \\
\cmidrule(lr){4-5}\cmidrule(lr){6-8}\cmidrule(l){9-11}
& & & $\mu_{\phi}$ & $\sigma_{\phi}$ & $\mu_{\phi}$ & $\sigma_{\phi}$ & $d$ & $\mu_{\phi}$ & $\sigma_{\phi}$ & $d$ \\
\midrule
\multirow{8}[2]{*}{IMDb} & \multirow{4}{*}{CNN} & \textsc{Random} & 7.6 & 2.5 & 3.5 & 2.8 & \opt{1.6} & 4.4 & 2.4  & 1.3 \\
 &  & \textsc{Prioritized} & 7.6 & 2.5 & 3.3 & 2.7 & \opt{1.6} & 3.9 & 2.5  & 1.5 \\
 &  & \textsc{Genetic} & 6.3 & 2.0 & 3.5 & 2.2 & 1.3 & 3.7 & 2.1 & 1.3 \\
 &  & \textsc{PWWS} & 6.7 & 2.3 & 4.0 & 2.4 & 1.1 & 4.5 & 2.1 & 1.0  \\ \cmidrule{2-11}
 & \multirow{4}{*}{LSTM} & \textsc{Random} & 7.6 & 2.5 & 3.5 & 2.8 & 1.6 & 4.4 & 2.4 & 1.3  \\
 &  & \textsc{Prioritized} & 7.6 & 2.5 & 2.8 & 2.3 & \opt{2.0} & 3.2 & 2.2 & 1.8  \\
 &  & \textsc{Genetic} & 6.2 & 2.0 & 3.1 & 1.9 & 1.6 & 3.3 & 1.8 & 1.5  \\
 &  & \textsc{PWWS} & 6.4 & 2.2 & 3.5 & 2.1 & 1.4 & 3.7 & 1.9 & 1.3  \\  \midrule
\multirow{8}[2]{*}{SST-2} & \multirow{4}{*}{CNN} & \textsc{Random} & 5.4 & 2.5 & 2.0 & 2.3 & \opt{1.4} & 3.8 & 1.7 & 0.7 \\
 &  & \textsc{Prioritized} & 5.4 & 2.5 & 2.4 & 2.1 & 1.3 & 3.5 & 1.7 & 0.8 \\
 &  & \textsc{Genetic} & 4.3 & 1.8 & 2.2 & 1.9 & 1.2 & 3.2 & 1.4 & 0.6 \\
 &  & \textsc{PWWS} & 4.8 & 2.1 & 2.8 & 2.1 & 1.0 & 3.8 & 1.5 & 0.6 \\ \cmidrule{2-11}
 & \multirow{4}{*}{LSTM} & \textsc{Random} &  5.4 & 2.6 & 2.0 & 2.3 & \opt{1.4} & 3.8 & 1.7 & 0.7 \\
 &  & \textsc{Prioritized} & 5.4 & 2.5 & 2.3 & 2.1 & 1.3 & 3.4 & 1.6 & 0.9 \\
 &  & \textsc{Genetic} & 4.3 & 1.7 & 2.0 & 1.9 & 1.3 & 3.1 & 1.4 & 0.8  \\
 &  & \textsc{PWWS} & 4.8 & 2.0 & 2.7 & 2.1 & 1.0 & 3.7 & 1.4 & 0.6 \\ \bottomrule
\end{tabular}
}
\caption{Mean $\log_e$ frequencies of replaced words and their corresponding substitutions by attack, model and dataset. The shown values are the mean $\mu_{\phi}$ and standard deviation $\sigma_{\phi}$ of the $\log_e$ frequencies corresponding to each setting, and additionally the Cohen's $d$ effect sizes for the substitutions. Values in bold denote largest effect sizes per dataset and model.}
\label{tab:statistical-results_no_transformer}
\end{table*}

\section{Dataset statistics}
\label{sec:app-datasets}
The SST-2 dataset comes with a pre-defined split of 67,349 samples for training, 872 for validation and 1,821 for testing. The IMDb dataset consists of 50,000 positive and negative movie reviews with a pre-defined split of 25,000 training and 25,000 test samples. Since this dataset does not have a pre-defined validation set, we hold out 1,000 randomly selected training set samples for validation. We select a validation set of roughly the same size as for SST-2 for fair comparisons when tuning parameters for adversarial example detection. To the best of our knowledge, the compared work~\cite{alzantot-etal-2018-generating, ren-etal-2019-generating} does not validate model performance on held-out training data.

\begin{table*}[t]
\centering
\resizebox{0.6\textwidth}{!}{
\begin{tabular}{cclcc}
\toprule
\textbf{Dataset} & \textbf{Model} & \multicolumn{1}{c}{\textbf{Attack}} & \multicolumn{1}{c}{\textbf{Subst.}} & 
\multicolumn{1}{c}{\textbf{non-OOV}} \\
\midrule
\multirow{12}[3]{*}{IMDb} & \multirow{4}{*}{CNN} & \textsc{Random} & $>10^{10594}$ & $>10^{7004}$ \\
 &  & \textsc{Prioritized} & $>10^{6549}$ & $>10^{5009}$ \\
 &  & \textsc{Genetic} & $>10^{2581}$ & $>10^{2318}$ \\
 &  & \textsc{PWWS} & $>10^{2182}$ & $>10^{1673}$ \\ \cmidrule{2-5}
 & \multirow{4}{*}{LSTM} & \textsc{Random} & $>10^{9643}$ & $>10^{6338}$ \\
 &  & \textsc{Prioritized} & $>10^{5949}$ & $>10^{4967}$ \\
 &  & \textsc{Genetic} & $>10^{2550}$ & $>10^{2369}$ \\
 &  & \textsc{PWWS} & $>10^{1666}$ & $>10^{1442}$ \\ \cmidrule{2-5}
 & \multirow{4}{*}{RoBERTa} & \textsc{Random} &  $>10^{12138}$ & $>10^{7948}$ \\
 &  & \textsc{Prioritized} & $>10^{9014}$ & $>10^{6043}$\\
 &  & \textsc{Genetic} & $>10^{4215}$ & $>10^{3672}$ \\
 &  & \textsc{PWWS} & $>10^{5182}$ & $>10^{3656}$ \\ \midrule
\multirow{12}[3]{*}{SST-2} & \multirow{4}{*}{CNN} & \textsc{Random} & $>10^{754}$ & $> 10^{138}$ \\
 &  & \textsc{Prioritized} & $>10^{573}$ & $>10^{222}$ \\
 &  & \textsc{Genetic} & $>10^{388}$ & $>10^{104}$ \\
 &  & \textsc{PWWS} & $>10^{397}$ & $>10^{131}$ \\ \cmidrule{2-5}
 & \multirow{4}{*}{LSTM} & \textsc{Random} & $>10^{800}$ & $> 10^{153}$\\
 &  & \textsc{Prioritized} & $>10^{648}$ & $>10^{264}$ \\
 &  & \textsc{Genetic} & $>10^{522}$ & $>10^{148}$ \\
 &  & \textsc{PWWS} & $>10^{456}$ & $>10^{144}$ \\ \cmidrule{2-5}
 & \multirow{4}{*}{RoBERTa} & \textsc{Random} & $>10^{867}$ & $>10^{149}$ \\
 &  & \textsc{Prioritized} & $>10^{779}$ & $>10^{130}$ \\
 &  & \textsc{Genetic} & $>10^{584}$ & $>10^{55}$ \\
 &  & \textsc{PWWS} & $>10^{600}$ & $>10^{125}$ \\
 \bottomrule
\end{tabular}
}
\caption{Bayes factors ($\mathrm{BF}_{10}$) for the Bayesian hypothesis tests.}
\label{tab:bayes-factors}
\end{table*}

\section{Model and attack details}
\label{sec:app-attack-details}

\subsection{RoBERTa}
We utilize a pre-trained RoBERTa (base) model~\cite{liu2019roberta} provided by the \textit{Hugging Face Transformers} library~\cite{Wolf2019HuggingFacesTS}. We use maximum input sequence lengths of 256 and 128 after byte-pair encoding~\cite{sennrich-etal-2016-neural} for the IMDb and SST-2 datasets, respectively. The RoBERTa model consists of 125 million parameters.\footnote{\url{https://github.com/pytorch/fairseq/tree/master/examples/roberta}} The model was trained for $10$ epochs with batch size $32$ (SST-2) and $16$ (IMDb) and a learning rate of $1 \cdot 10^{-5}$. We evaluated model performance after each epoch on the validation set and selected the best-performing checkpoints for testing.

\subsection{CNN/LSTM} 
The CNN architecture consists of $3$ convolutional layers with kernel sizes $2$, $3$ and $4$ and $100$ feature maps for each convolutional layer. The LSTM operates on a hidden state size of $128$. Following~\citet{alzantot-etal-2018-generating}, we initialize the LSTM with pre-trained \textsc{GloVe}~\cite{pennington-etal-2014-glove} word embeddings, and do the same for the CNN.

Both the LSTM and the CNN use \textit{Dropout}~\cite{JMLR:v15:srivastava14a} during training with a rate of $0.1$ before applying the output layer. We trained both models for $20$ epochs using the \textit{Adam} optimizer~\citep{kingma2014}. We evaluated model performance after each epoch on the validation set and selected the best-performing checkpoints for testing. The CNN and LSTM models were trained with batch size $100$ and a learning rate of $1 \cdot 10^{-3}$.

\subsection{PWWS}
Our implementation of \textsc{PWWS} is based on the code as provided by~\citet{ren-etal-2019-generating} on GitHub.\footnote{\url{https://github.com/JHL-HUST/PWWS}}

\subsection{\textsc{Genetic}}
Note that we utilize a different language model for the \texttt{Perturb} subroutine as compared to the original implementation by~\citet{alzantot-etal-2018-generating}. While~\citet{alzantot-etal-2018-generating} employ the Google $1$ billion words language model~\cite{41880}, we instead utilize the recently proposed GPT-2 language model~\cite{radford2019language} and compute the sequences' perplexity scores using the exponentialized language modelling loss (we employ the pre-trained \texttt{GPT2LMHeadModel} language model from~\citet{Wolf2019HuggingFacesTS}). We compute the perplexity scores for each perturbed sequence only around the respective replacement words by only considering a subsequence ranging from five words before to five words after an inserted replacement. The motivation for using a different language model as compared to the original implementation is due to computational efficiency, since we observed a notable decrease in attack runtime with our modification. This does not have an impact on attack performance, since our implementation of the \textsc{Genetic} has an attack success rate of 98.6\% against the LSTM on IMDb, whereas~\citet{alzantot-etal-2018-generating} report an attack success rate of 97\%.

For attacks against SST-2, we furthermore increase the $\delta$ threshold for the maximum distance between replaced words and substitutions to $\delta = 1.0$, since we observed poor attack performances with $\delta = 0.5$ (which was used by~\citet{alzantot-etal-2018-generating} and in our experiments on IMDb). All other parameters of the attack (e.g., the number of generations and population size) are directly adapted from~\citet{alzantot-etal-2018-generating}.

We restrict the words eligible for replacement by the \textsc{Genetic} attack to non-stopwords, in accordance to~\citet{alzantot-etal-2018-generating}. Since the attack computes nearest neighbors for a selected word from a pre-trained embedding space, we furthermore can only select words for which there exists an embedding representation in this pre-trained space. On the SST-2 test set, we found three input sequences consisting of only one word which we excluded from our evaluation, since the used GPT-2 language model implementation requires an input sequence consisting of more than one word.

\begin{figure*}[t]
\resizebox{1.0\textwidth}{!}{
    \subfloat[\textsc{Random} on SST-2]{{\includegraphics[width=0.5\textwidth]{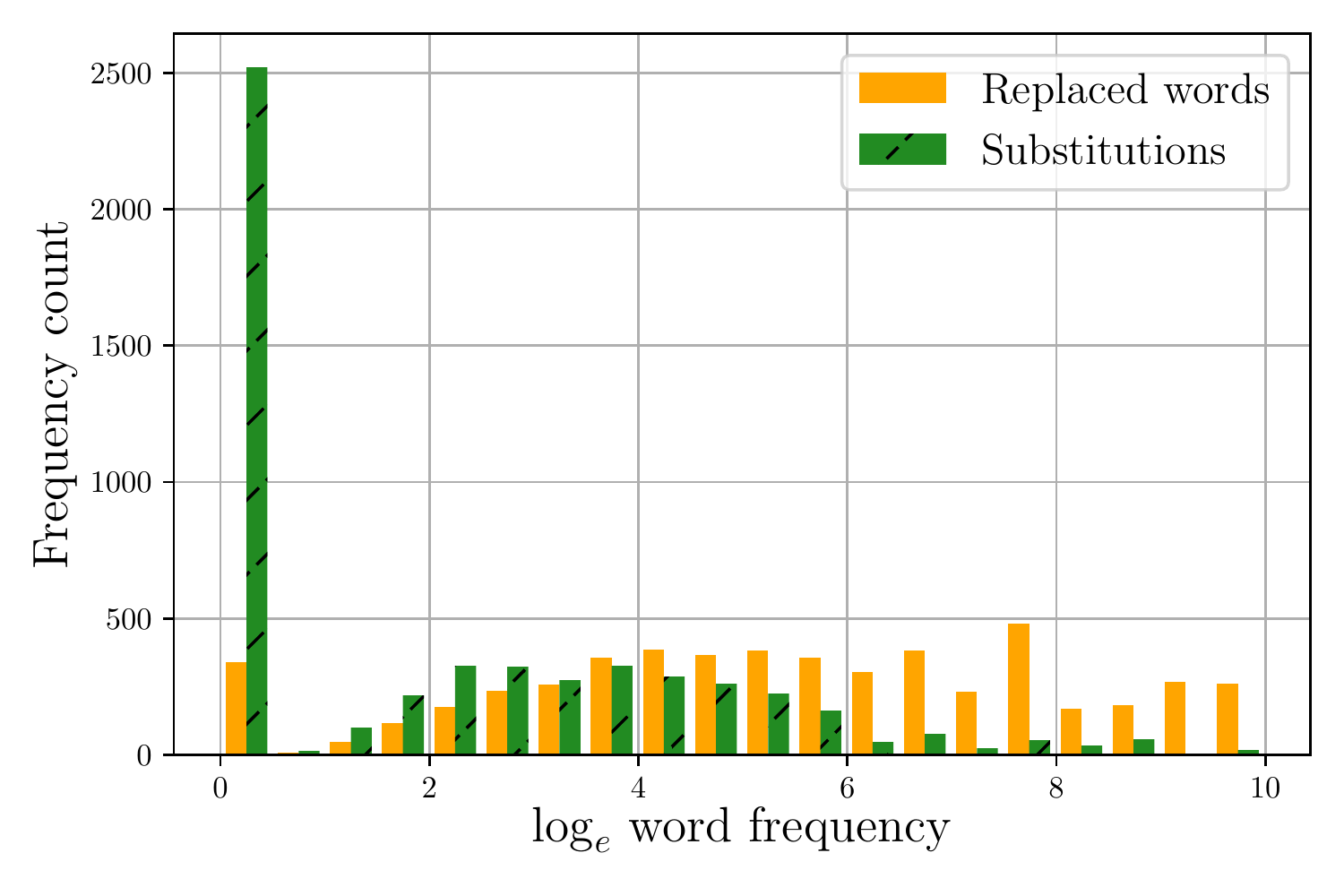}}}
    \subfloat[\textsc{Random} on IMDb]{{\includegraphics[width=0.5\textwidth]{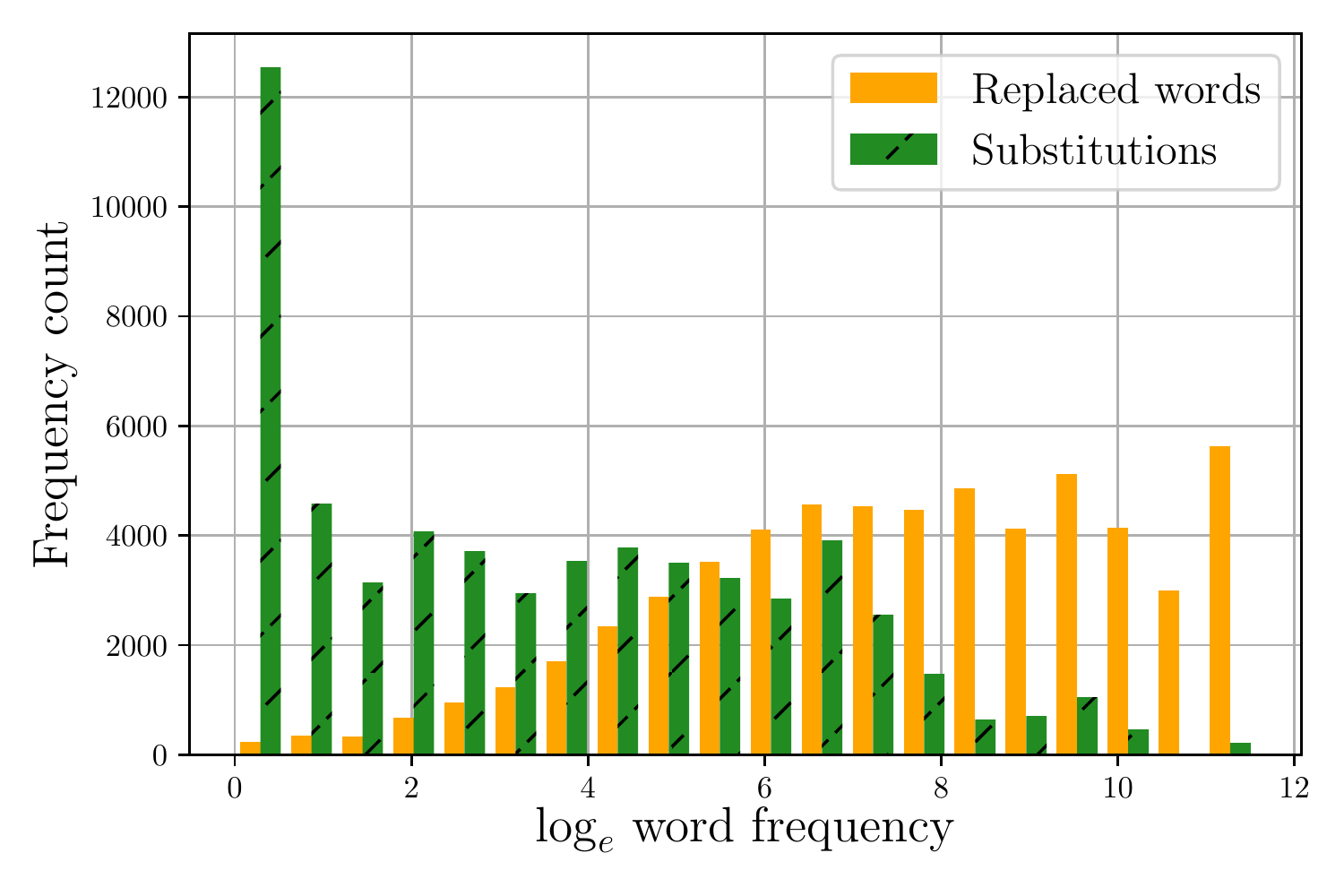}}}
}
\resizebox{1.0\textwidth}{!}{
    \subfloat[\textsc{Prioritized} on SST-2]{{\includegraphics[width=0.5\textwidth]{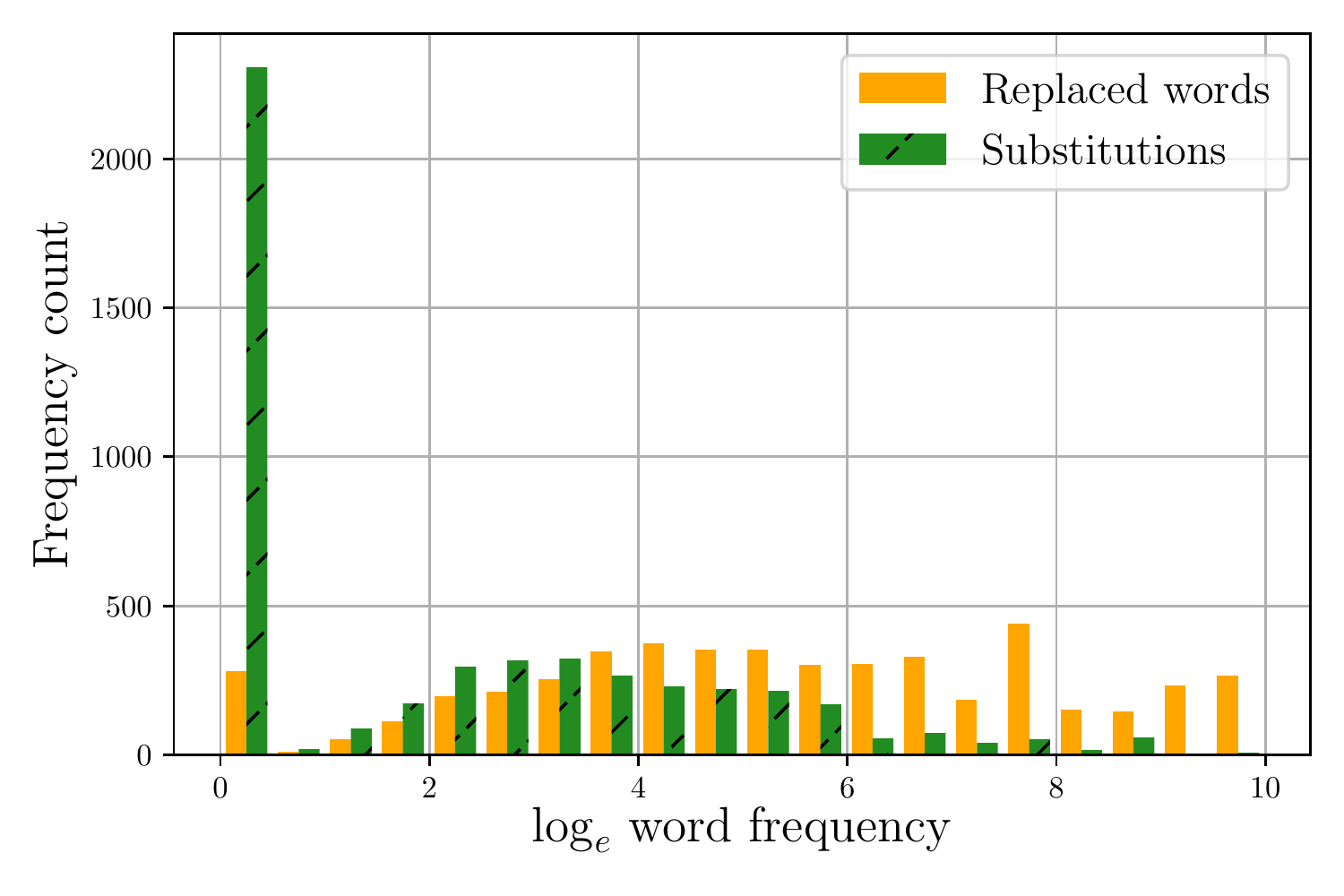}}}
    \subfloat[\textsc{Prioritized} on IMDb]{{\includegraphics[width=0.5\textwidth]{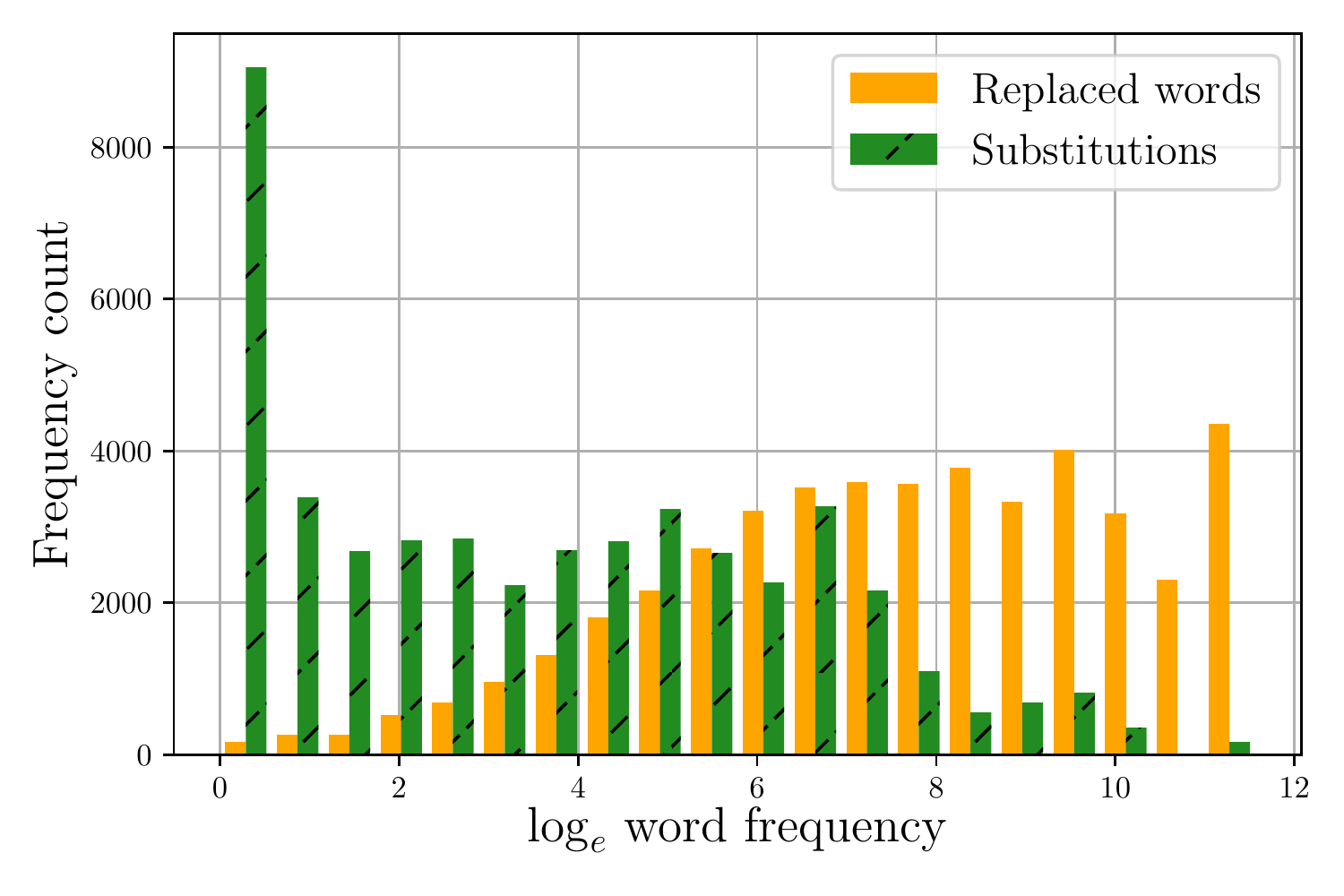}}}
}
\resizebox{1.0\textwidth}{!}{
    \subfloat[\textsc{Genetic} on SST-2]{{\includegraphics[width=0.5\textwidth]{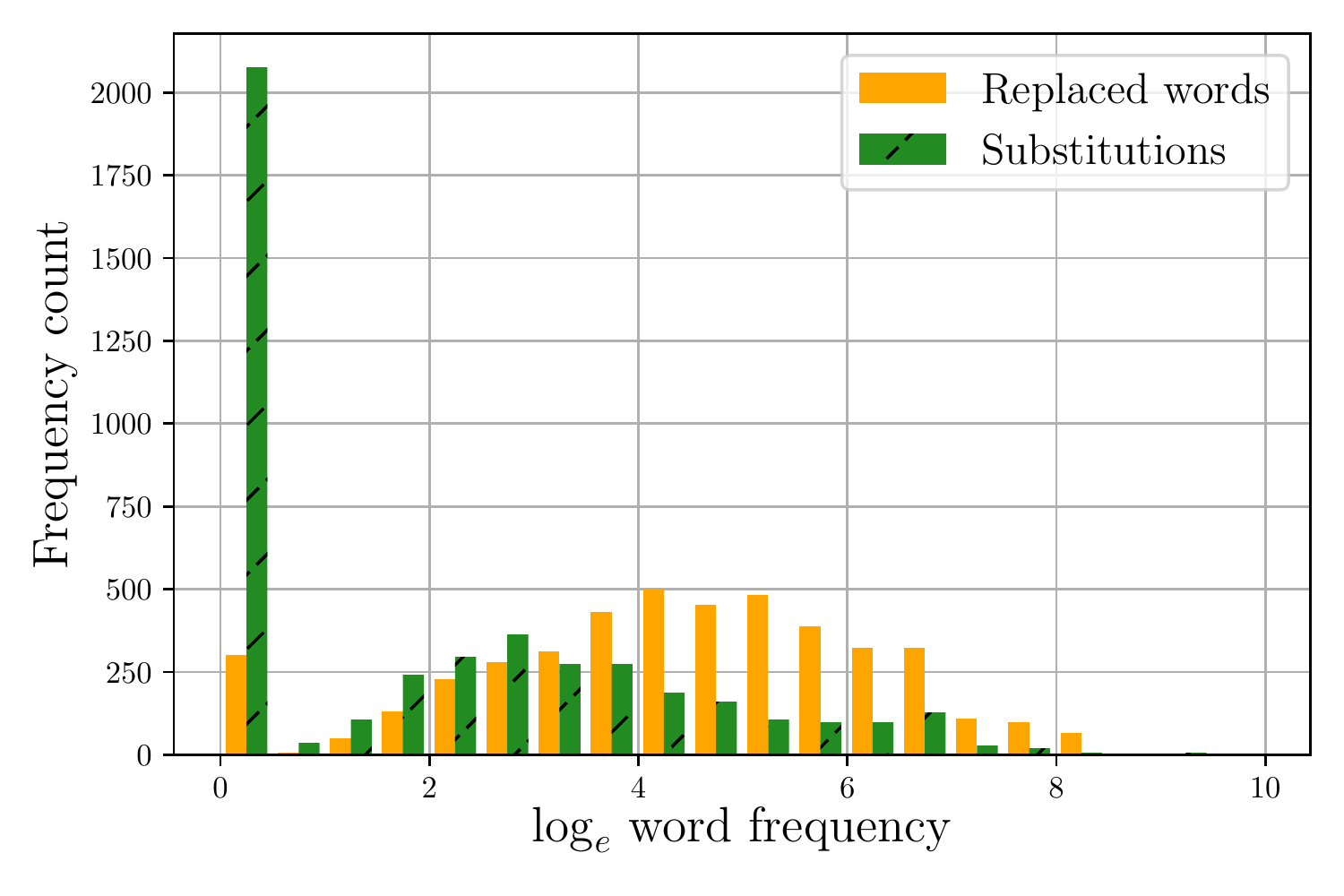}}}
    \subfloat[\textsc{Genetic} on IMDb]{{\includegraphics[width=0.5\textwidth]{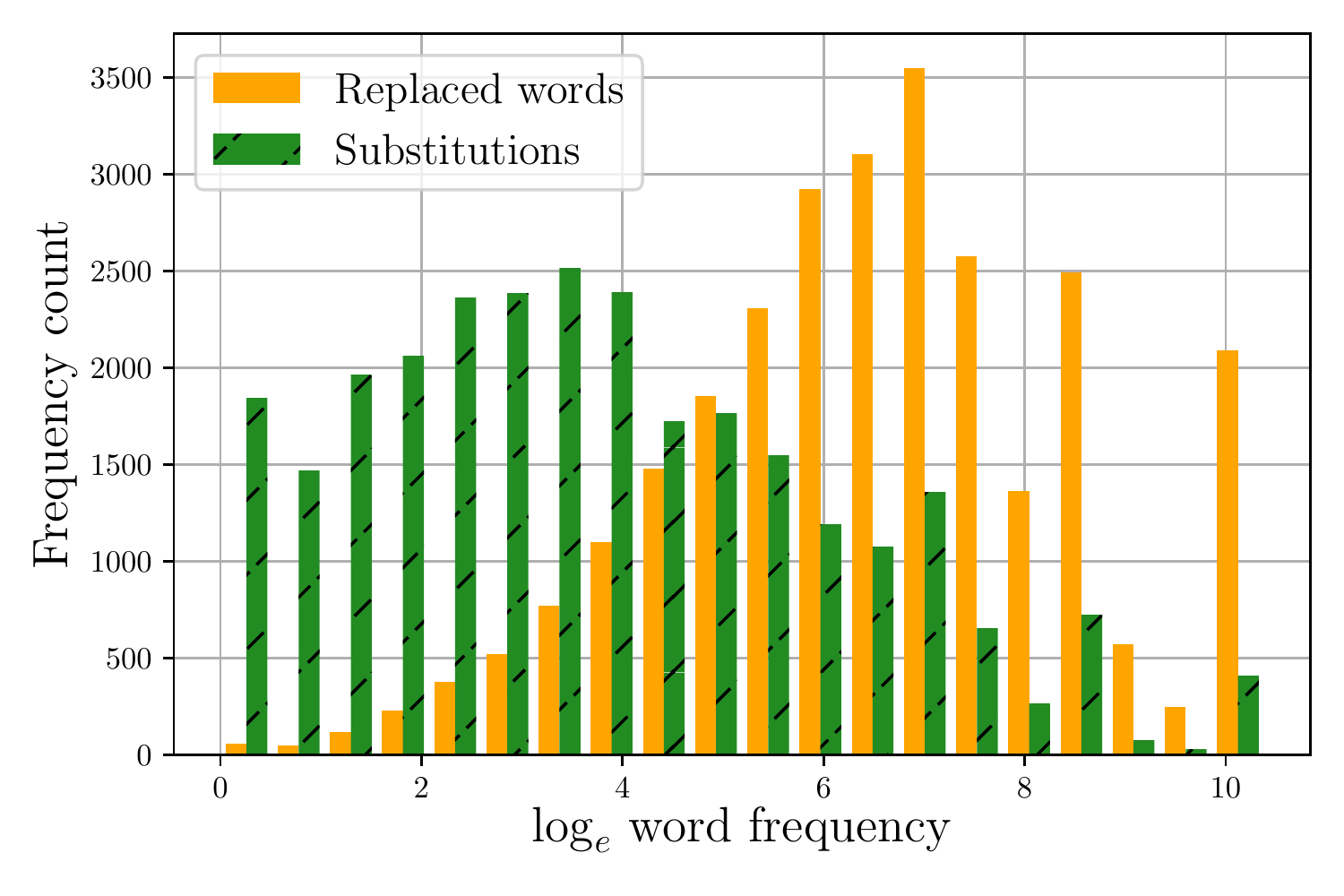}}}
}    
\resizebox{1.0\textwidth}{!}{
    \subfloat[\textsc{PWWS} on SST-2]{{\includegraphics[width=0.5\textwidth]{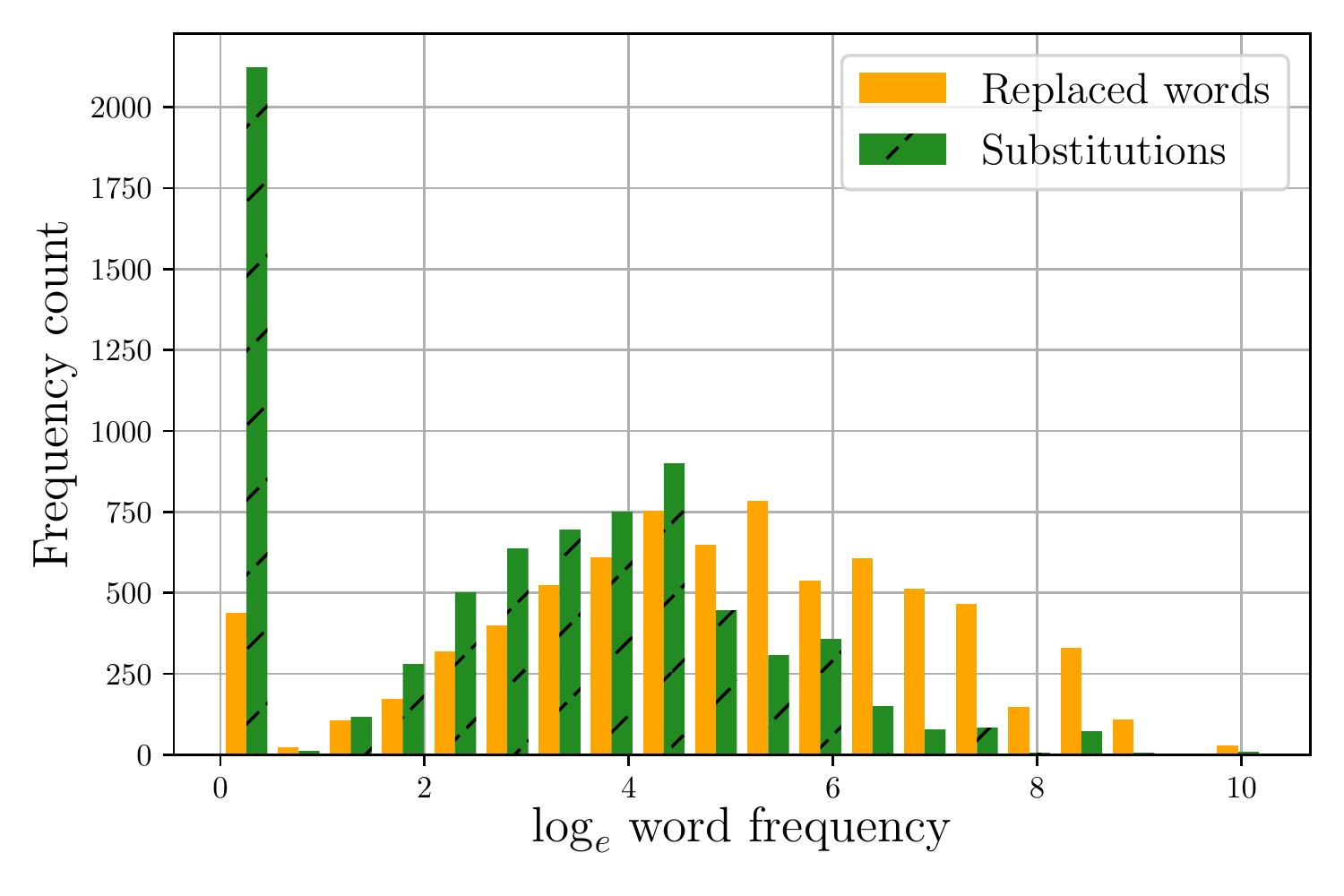}}}
    \subfloat[\textsc{PWWS} on IMDb]{{\includegraphics[width=0.5\textwidth]{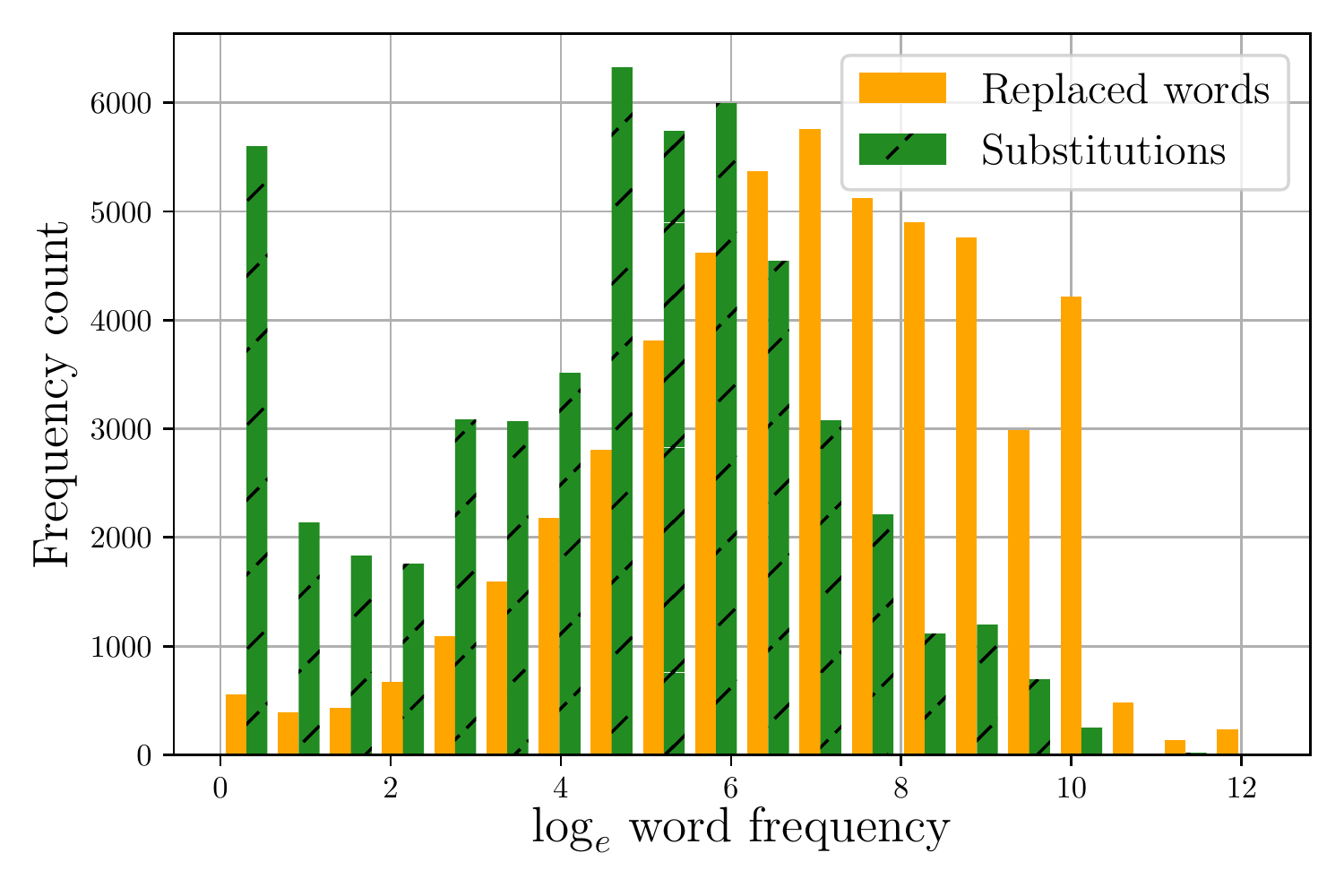}}}
}
\caption{Histograms showing the frequency distribution of words replaced by the attacks and their corresponding substitutions against the RoBERTa model. The $x$-axis represents the words' $\log_e$ frequency with respect to the model's training corpus, the $y$-axis denotes their respective frequencies among the perturbed test set sequences.}
\label{fig:frequency-difference-visualizations}
\end{figure*}

\subsection{\textsc{Random}, \textsc{Prioritized}, \textsc{PWWS}, \textsc{Genetic}}
For the \textsc{Genetic} attack, we follow~\citet{alzantot-etal-2018-generating} by limiting the maximum amount of word replacements to 20\% of the input sequence length. We apply the same threshold to the \textsc{Random} and \textsc{Prioritized} attacks, but not to \textsc{PWWS} since we observed low replacement rates despite the attack's effectiveness. This is in agreement to the results reported in~\citet{ren-etal-2019-generating}.

\section{Frequency differences for CNN and LSTM models}
\label{sec:app-freq-diff-cnn-lstm}
The $\log_e$ frequencies for the four attacks against the CNN and LSTM models can be found in Table~\ref{tab:statistical-results_no_transformer}. In accordance to the experiments with RoBERTa (see Section~\ref{sec:statistical-analysis} in the paper), we observe large Cohen's $d$ effect sizes for the majority of the comparisons, which shows that the statistical frequency differences between replaced words and their substitutions are present for adversarial attacks against these two models as well. 

\section{Bayes factors}
\label{sec:app-bayes-factor}
The Bayes factors for the mean frequency comparisons between replaced words and their adversarial substitutions can be found in Table~\ref{tab:bayes-factors}. We observe high values for $\mathrm{BF}_{10}$ across all comparisons, providing strong evidence for the hypothesis that the $\log_e$ frequency means between replaced words and their substitutions are different.

\begin{figure*}[t]
\centering
\resizebox{1.0\textwidth}{!}{
    \subfloat[IMDb]{{\includegraphics[width=0.5\textwidth]{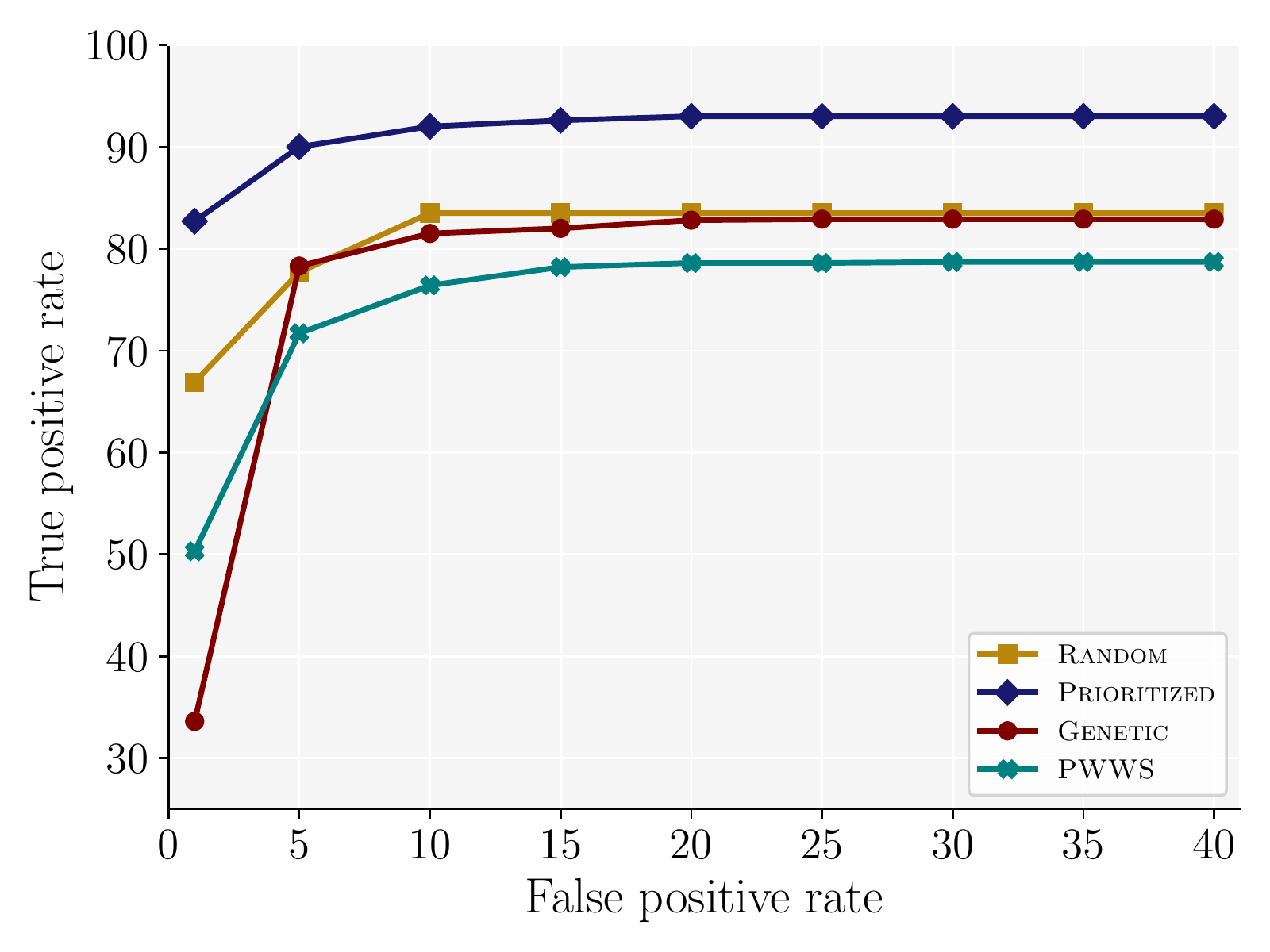}}}
    \qquad
    \subfloat[SST-2]{{\includegraphics[width=0.5\textwidth]{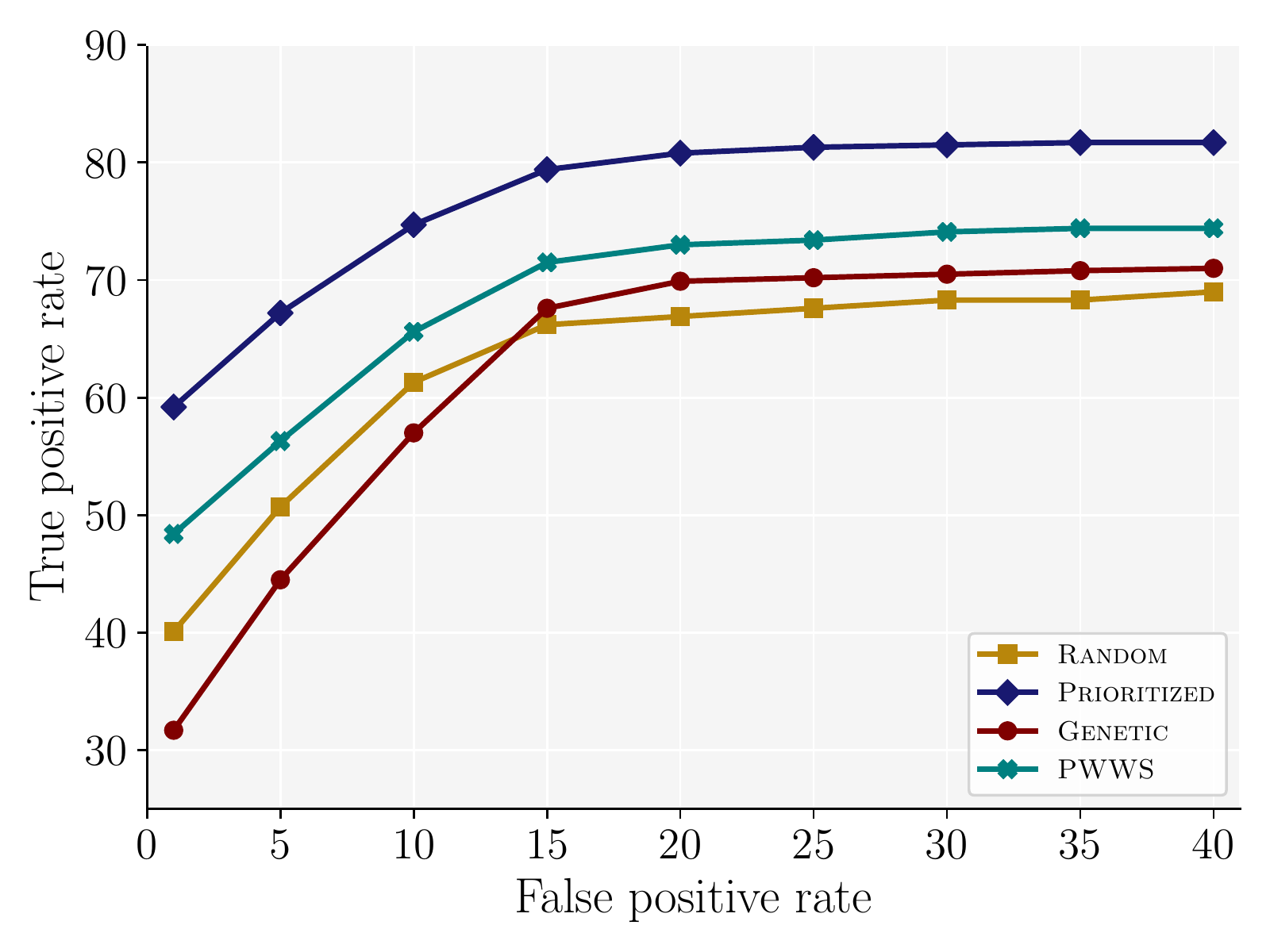}}}
}
\caption{The trade-off between true and false positive rates on the test sets with all four attacks against RoBERTa on (a) IMDb and (b) SST-2. The true positive rates ($y$-axis) are computed when $\gamma$ is set to allow for different quasi-fixed amounts of false positives ($x$-axis).}
\label{fig:delta-performance-thresholds-detection}
\end{figure*}

\section{Visualizations of frequency differences}
\label{sec:app-freq-histograms}
Figure~\ref{fig:frequency-difference-visualizations} illustrates the frequency differences for attacks against the RoBERTa model using histograms. We observe that for the majority of the attacks, OOV substitutions occur most often among the perturbed sequences.

\section{Varying false positive thresholds}
\label{sec:app-varying-fpr}
The rate of false positives predicted by a detection system is crucial for its practicability, and a limited amount of false positives is hence highly desirable. Figure~\ref{fig:delta-performance-thresholds-detection} illustrates the true positive rates predicted by \textsc{FGWS} for all attacks against RoBERTa with different quasi-fixed false positive thresholds (as in Section~\ref{subsec:experiments} of the paper, $\delta$ was tuned on the validation set for each value of $\gamma$ corresponding to the specific false positive threshold). As expected, we observe a trade-off between true and false positive rates for varying values of $\gamma$, such that lower false positive rates imply lower true positive rates. However, even for false positive rates of 1\% and 5\%, we observe that \textsc{FGWS} is able to detect between 33.6\% and 90.0\% of adversarial examples on IMDb and between 31.7\% and 67.2\% on SST-2. This indicates that \textsc{FGWS} has the potential to detect a useful fraction of adversarial examples without creating an excessive burden of false positives. 

\section{Additional FGWS examples}
\label{sec:app-fgws-examples}
Additional examples of \textsc{FGWS} can be found in Table~\ref{tab:app-tp} (true positives), Table~\ref{tab:app-fp} (false positives), Table~\ref{tab:app-tn} (true negatives) and Table~\ref{tab:app-fn} (false negatives).

\begin{table*}[t]
\small 

\begin{center}
\textbf{Model}:RoBERTa on SST-2
\end{center}

\begin{tabularx}{\textwidth}{lXc}
\toprule 
Unperturbed & {first good then bothersome} & \textit{negative} (74.5\%) \\ \midrule
\textsc{Genetic} &  first good then \advsubstnoc{0.00}{galling}{0.00}{bothersome} & \textit{positive} (88.7\%)\\ \midrule
\textsc{DISP} &  first good \advsubstnoc{8.96}{that}{5.31}{then} galling & \textit{positive} (84.8\%)\\ \midrule
\textsc{FGWS} &  first good then \advsubstnoc{4.32}{annoying}{0.00}{galling} & \textit{negative} (91.3\%)\\ \bottomrule
\end{tabularx}

\vskip 0.8cm

\begin{center}
\textbf{Model}:RoBERTa on IMDb
\end{center}

\begin{tabularx}{\textwidth}{lXc}
\toprule
Unperturbed & {i am a huge rupert everett fan . i adore kathy bates so when i saw it available i decided to check it out . the synopsis didn t really tell you much . in parts it was silly touching and in others some parts were down right hysterical . any person that is a huge fan of a personality of any type will find some small identifying traits with the main character . of course there are many they won t but that is the point if you like any of the actors give it a watch but don t look for any thing too dramatic it s good fun . i might also mention you can see how darn tall rupert is . i mean i knew he was 6 4 but he seems even more in this film . he even seemed to stoop a bit due to the other characters height in this . he is tall i mean tall and for you rupert fans there is a bare chest scene ... wonderful} & \textit{positive} (99.2\%) \\ \midrule
\textsc{PWWS} &  i am a huge rupert everett fan . i adore kathy bates so when i saw it available i decided to \advsubstnoc{6.54}{stop}{6.19}{check} it out . the synopsis didn t really tell you much . in parts it was silly touching and in others some parts were down right hysterical . any person that is a huge fan of a personality of any type will find some small identifying traits with the main character . of course there are many they won t but that is the point if you like any of the actors give it a watch but don t look for any thing too dramatic it s \advsubstnoc{0.00}{undecomposed}{9.22}{good} fun . i might also mention you can see how darn tall rupert is . i mean i knew he was 6 4 but he seems even more in this film . he even seemed to stoop a bit \advsubstnoc{0.00}{imputable}{6.31}{due} to the other characters height in this . he is tall i mean tall and for you rupert fans there is a bare chest scene ... \advsubstnoc{4.45}{tremendous}{7.08}{wonderful} & \textit{negative} (60.1\%)\\ \midrule
\textsc{DISP} &  i am a huge rupert everett fan . i adore kathy bates so when i saw it available i decided to \advsubstnoc{9.27}{out}{6.54}{stop} it out . the synopsis didn t really tell you much . in parts it was silly touching and in others some parts were down right hysterical . any person that is a huge fan of a personality of any type will find some small identifying traits with the main character . of course there are many they won \advsubstnoc{0.00}{,}{9.97}{t} but that is the point if you like any of the actors give it a watch but don t look for any thing too dramatic it \advsubstnoc{0.00}{'s}{10.54}{s} \advsubstnoc{9.50}{so}{0.00}{undecomposed} fun . i \advsubstnoc{9.20}{can}{7.44}{might} also mention you can see how darn tall rupert is . i mean i knew he was 6 4 but he seems even more in this film . he even seemed to stoop a bit imputable to the other characters height in this . he is tall i mean tall and for you rupert fans there is a bare chest scene ... \advsubstnoc{12.05}{.}{4.45}{tremendous} & \textit{positive} (92.0\%)\\ \midrule
\textsc{FGWS} &  i am a huge rupert everett fan . i adore kathy bates so when i saw it available i decided to stop it out . the synopsis didn t really tell you much . in parts it was silly touching and in others some parts were down right hysterical . any person that is a huge fan of a personality of any type will find some small \advsubstnoc{7.26}{place}{2.48}{identifying} traits with the main character . of course there are many they won t but that is the point if you like any of the actors give it a watch but don t look for any thing too dramatic it s \advsubstnoc{9.22}{good}{0.00}{undecomposed} fun . i might also mention you can see how darn tall rupert is . i mean i knew he was 6 4 but he seems even more in this film . he even seemed to \advsubstnoc{6.22}{sit}{2.40}{stoop} a bit \advsubstnoc{6.31}{due}{0.00}{imputable} to the other characters height in this . he is tall i mean tall and for you rupert fans there is a bare chest scene ... tremendous & \textit{positive} (88.9\%)\\ \bottomrule
\end{tabularx}

\caption{Illustration of true positives generated with \textsc{FGWS} against RoBERTa on SST-2 (top) and IMDb (bottom). The substitutions caused the model to change the predicted label back to its ground-truth for the given adversarial examples.}
\label{tab:app-tp}
\end{table*}

\begin{table*}[t]
\small
\begin{center}
    \textbf{Model}: RoBERTa on SST-2
\end{center}
\begin{tabularx}{\textwidth}{lXc}
\toprule
Unperturbed & {imagine if you will a tony hawk skating video interspliced with footage from behind enemy lines and set to jersey shore techno} & \textit{negative} (83.6\%) \\ \midrule
\textsc{DISP} &  imagine if you \advsubstnoc{5.97}{get}{6.84}{will} a tony hawk skating video \advsubstnoc{0.00}{,}{0.00}{interspliced} with footage from behind enemy lines and set to jersey shore techno & \textit{negative} (87.1\%)\\ \midrule
\textsc{FGWS} &  imagine if you will a \advsubstnoc{3.76}{kevin}{1.10}{tony} \advsubstnoc{3.93}{pitch}{2.48}{hawk} skating video interspliced with footage from behind enemy lines and set to \advsubstnoc{6.55}{new}{2.08}{jersey} \advsubstnoc{3.93}{sea}{2.08}{shore} \advsubstnoc{5.69}{music}{1.61}{techno} & \textit{positive} (65.7\%)\\ \bottomrule
\end{tabularx}

\vskip 0.8cm

\begin{center}
    \textbf{Model}: RoBERTa on IMDb
\end{center}
\small
\begin{tabularx}{\textwidth}{lXr}
\toprule
Unperturbed & admittedly alex has become a little podgey but they are still for me the greatest rock trio ever . i wholeheartedly recommend this dvd to any fan . i was very disappointed that they canceled their planned recent munich gig logistics and regret not making an effort to see them elsewhere . the dvd is a small consolation the greatest incentive to acquire a proper dvd playback setup . naive perhaps but i still don t understand the significance of the tumble driers on stage i would be grateful for any clarification . cheers iain . & \textit{positive} (99.4\%) \\ \midrule
\textsc{DISP} &  admittedly alex has become a little podgey but they are still for me the greatest rock trio ever . i wholeheartedly recommend this dvd to any fan . i was very disappointed that they canceled their planned recent munich gig logistics and regret not making an effort to see them elsewhere . the dvd is a small consolation the greatest incentive to acquire a proper dvd playback setup . naive perhaps but i still don t understand the significance of the \advsubstnoc{9.77}{one}{1.95}{tumble} driers on stage i would be grateful for any clarification . cheers \advsubstnoc{10.73}{that}{0.00}{iain} . & \textit{positive} (99.3\%)\\ \midrule
\textsc{FGWS} &  admittedly alex has become a little podgey but they are still for me the greatest rock trio ever . i \advsubstnoc{4.55}{disagree}{2.30}{wholeheartedly} recommend this dvd to any fan . i was very disappointed that they canceled their planned recent \advsubstnoc{5.03}{germany}{2.08}{munich} gig \advsubstnoc{3.40}{transport}{0.69}{logistics} and regret not making an effort to see them elsewhere . the dvd is a small \advsubstnoc{5.69}{win}{2.08}{consolation} the greatest \advsubstnoc{5.53}{opportunity}{1.61}{incentive} to acquire a proper dvd \advsubstnoc{6.21}{editing}{2.08}{playback} setup . naive perhaps but i still don t understand the significance of the \advsubstnoc{6.19}{fall}{1.95}{tumble} \advsubstnoc{1.61}{dryer}{0.00}{driers} on stage i would be grateful for any \advsubstnoc{5.11}{explanation}{1.10}{clarification} . cheers iain . & \textit{negative} (50.1\%)\\ \bottomrule
\end{tabularx}
\caption{Illustration of false positives generated with \textsc{FGWS} against RoBERTa on SST-2 (top) and IMDb (bottom). The substitutions caused the model to change the predicted label for the given unperturbed sequences.}
\label{tab:app-fp}
\end{table*}

\begin{table*}[t]
\small
\begin{center}
    \textbf{Model}: RoBERTa on SST-2
\end{center}
\begin{tabularx}{\textwidth}{lXc}
\toprule
Unperturbed & {it s a hoot and a half and a great way for the american people to see what a candidate is like when he s not giving the same 15 cent stump speech} & \textit{positive} (100.0\%) \\ \midrule
\textsc{DISP} &  it \advsubstnoc{0.00}{'s}{9.09}{s} a hoot and a half and a great way for the american people to see what a candidate is like when he \advsubstnoc{0.00}{'s}{9.09}{s} not giving the same 15 \advsubstnoc{6.01}{minutes}{0.00}{cent} \advsubstnoc{10.22}{the}{0.00}{stump} speech & \textit{positive} (100.0\%)\\ \midrule
\textsc{FGWS} &  it s a hoot and a half and a great way for the american people to see what a \advsubstnoc{3.71}{nomination}{1.95}{candidate} is like when he s not giving the same 15 cent \advsubstnoc{2.48}{stamp}{0.00}{stump} \advsubstnoc{4.45}{words}{0.00}{speech} & \textit{positive} (100.0\%)\\ \bottomrule
\end{tabularx}

\vskip 0.8cm

\begin{center}
    \textbf{Model}: RoBERTa on IMDb
\end{center}
\small
\begin{tabularx}{\textwidth}{lXr}
\toprule
Unperturbed & {it was awful plain and simple . what was their message where was the movie going with this it has all the ingredients of a sub b grade movie . from plotless storyline the bad acting to the cheesey slow mo cinematography . i d sooner watch a movie i ve already seen like goodfellas a bronx tale even grease . there are no likeable characters . in the end you just want everyone to die already . save 2 hours of your life and skip this one .} & \textit{negative} (99.9\%) \\ \midrule
\textsc{DISP} &  it was awful plain and simple . what was their message where was the movie going with this it has all the ingredients of a sub b grade movie . from plotless storyline the bad acting to the cheesey slow mo cinematography . i \advsubstnoc{8.94}{would}{7.56}{d} sooner watch a movie i \advsubstnoc{9.79}{have}{8.17}{ve} already seen like goodfellas a bronx tale \advsubstnoc{10.97}{in}{8.97}{even} grease . there are no likeable characters . in the end you just want everyone to die already . save 2 hours of your life and skip this one . & \textit{negative} (99.9\%)\\ \midrule
\textsc{FGWS} &  it was awful plain and simple . what was their message where was the movie going with this it has all the ingredients of a sub b grade movie . from \advsubstnoc{4.28}{unwatchable}{1.39}{plotless} storyline the bad acting to the \advsubstnoc{6.10}{cheesy}{1.95}{cheesey} slow mo cinematography . i d sooner watch a movie i ve already seen like goodfellas a bronx tale even grease . there are no likeable characters . in the end you just want everyone to die already . save 2 hours of your life and skip this one . & \textit{negative} (99.9\%)\\ \bottomrule
\end{tabularx}
\caption{Illustration of true negatives generated with \textsc{FGWS} against RoBERTa on SST-2 (top) and IMDb (bottom). The substitutions did not cause the model to change the predicted label for the given unperturbed sequences.}
\label{tab:app-tn}
\end{table*}

\begin{table*}[t]
\small
\begin{center}
    \textbf{Model}: RoBERTa on SST-2
\end{center}
\begin{tabularx}{\textwidth}{lXc}
\toprule
Unperturbed & {the spark of special anime magic here is unmistakable and hard to resist} & \textit{positive} (100.0\%) \\ \midrule
\textsc{PWWS} &  the spark of special anime \advsubstnoc{2.83}{deception}{4.52}{magic} here is unmistakable and \advsubstnoc{2.77}{laborious}{6.15}{hard} to \advsubstnoc{4.58}{hold}{3.91}{resist} & \textit{negative} (84.4\%)\\ \midrule
\textsc{DISP} &  the spark of special anime deception here is unmistakable and \advsubstnoc{4.88}{able}{2.77}{laborious} to hold & \textit{positive} (99.9\%)\\ \midrule
\textsc{FGWS} &  the spark of special anime deception here is \advsubstnoc{4.52}{subtle}{2.48}{unmistakable} and laborious to hold & \textit{negative} (97.8\%)\\ \bottomrule
\end{tabularx}

\vskip 0.8cm

\begin{center}
    \textbf{Model}: RoBERTa on IMDb
\end{center}
\small
\begin{tabularx}{\textwidth}{lXr}
\toprule
Unperturbed & {graduation day is a result of the success of friday the 13th . both of those films are about creative bloody murders rather than suspense . if you enjoy that type of film i d recommend graduation day . if not i wouldn t. there s nothing new here just the same old killings . even though i ve given the film a 4 out of 10 i will say that it s not a repulsive film . it is watchable if your curious about it just not creative .} & \textit{negative} (71.3\%) \\ \midrule
\textsc{Genetic} &  graduation day is a result of the success of friday the 13th . both of those films are about creative bloody murders rather than suspense . if you enjoy that type of film i d recommend graduation day . if not i wouldn t. there s nothing new here just the same \advsubstnoc{5.06}{ancient}{8.03}{old} killings . even though i ve given the film a 4 out of 10 i will say that it s not a repulsive film . it is watchable if your curious about it just not creative . & \textit{positive} (53.5\%)\\ \midrule
\textsc{DISP} &  graduation day is a result of the success of friday the 13th . both of those films are about creative bloody murders rather than suspense . if you enjoy that type of film i \advsubstnoc{8.94}{would}{7.56}{d} recommend graduation day . if not i \advsubstnoc{8.64}{do}{6.48}{wouldn} t. there \advsubstnoc{11.14}{is}{10.54}{s} nothing new here just the same ancient killings . even though i \advsubstnoc{9.79}{have}{8.17}{ve} given the film a 4 out of 10 i will say that it \advsubstnoc{0.00}{'s}{10.54}{s} not a \advsubstnoc{9.22}{good}{3.69}{repulsive} film . it is watchable if your curious about it \advsubstnoc{11.14}{is}{9.32}{just} not creative . & \textit{negative} (99.5\%)\\ \midrule
\textsc{FGWS} &  graduation day is a result of the success of friday the 13th . both of those films are about creative bloody murders rather than suspense . if you enjoy that type of film i d recommend graduation day . if not i wouldn t. there s nothing new here just the same ancient killings . even though i ve given the film a 4 out of 10 i will say that it s not a repulsive film . it is watchable if your curious about it just not creative . & \textit{positive} (53.5\%)\\ \bottomrule
\end{tabularx}
\caption{Illustration of false negatives generated with \textsc{FGWS} against RoBERTa on SST-2 (top) and IMDb (bottom). The substitutions did not cause the model to change the predicted label back to its ground-truth for the given adversarial examples.}
\label{tab:app-fn}
\end{table*}
\end{document}